\documentclass[conference]{IEEEtran}
\IEEEoverridecommandlockouts
\usepackage{stfloats}
\usepackage{multirow}
\usepackage{amsmath,amssymb,amsfonts}
\usepackage{graphicx}
\usepackage{textcomp}
\usepackage{xcolor}
\usepackage{booktabs,tabularx,array,pifont}
\usepackage{cite}

\graphicspath{{figures/}}

\newcolumntype{L}[1]{>{\raggedright\arraybackslash}p{#1}}
\newcolumntype{C}[1]{>{\centering\arraybackslash}p{#1}}
\newcommand{\best}[1]{\textbf{#1}}
\newcommand{\model}[1]{\texttt{#1}}
\newcommand{\cmark}{\ding{51}}
\newcommand{\na}{--}
\newcommand{\unclear}{\ensuremath{?}}
\newcommand{\ci}[2]{~[#1--#2]}
\newcommand{\MAE}{\operatorname{MAE}}

\def\BibTeX{{\rm B\kern-.05em{\sc i\kern-.025em b}\kern-.08em
    T\kern-.1667em\lower.7ex\hbox{E}\kern-.125emX}}

\begin{document}

\title{Behind-the-Meter Photovoltaic Visibility from Net Smart Meters: Deployment Regimes and Operational Insights}

\author{\IEEEauthorblockN{Jack Zheng\textsuperscript{1}, Hao Wang\textsuperscript{1,2*}}
\IEEEauthorblockA{\textsuperscript{1}Department of Data Science and AI, Faculty of IT, Monash University, Australia \\
\textsuperscript{2}Monash Energy Institute, Monash University, Australia\\
\{jack.zheng, hao.wang2\}@monash.edu
}
\thanks{*Corresponding author: Hao Wang.}
\thanks{This work was supported in part by the Australian Research Council (ARC) Discovery Early Career Researcher Award (DECRA) under Grant DE230100046.}
}

\maketitle

\begin{abstract}
Distribution operators need visibility of behind-the-meter photovoltaic (PV) generation, yet most residential customers are observed only through net smart meters. Deep-learning-based PV disaggregation can restore this visibility, but its operational value depends on whether the model serves known customers, new customers, future periods, or new regions. We quantify this dependence using four deployment regimes that separate household transfer from temporal extrapolation, applied to Ausgrid smart-meter data with cross-dataset adaptation to the Pecan Street data. Using a fixed training protocol, we benchmark representative deep sequence architectures for PV disaggregation under regime-matched evaluation. At a fixed 48-hour context length, daylight mean absolute error (MAE) of the best model rises from 0.104~kW on seen households in the future test period to 0.185~kW on unseen households over the same period, and under a fixed training contract the deployment regime explains far more of the error variation than model choice. Under target-domain normalization, fine-tuning on five target-region households reduces source-model transfer error by 28\%--36\%. Long input windows help known-customer extrapolation, while their new-customer benefits vary by model. These findings suggest regime-matched error budgets, limited target supervision for model adaptation in new regions, and use-case-dependent context selection for energy data analytics.
\end{abstract}

\begin{IEEEkeywords}
Behind-the-meter photovoltaic, smart meter, distribution grid visibility, energy data analytics, deployment regimes, transfer learning.
\end{IEEEkeywords}

\section{Introduction}

\begin{figure*}[t]
\centering
\includegraphics[width=0.80\textwidth]{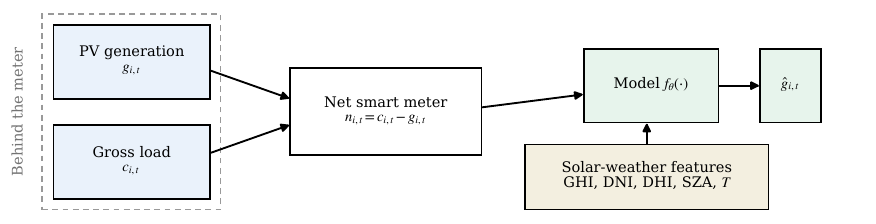}
\caption{Behind-the-meter PV disaggregation problem. The smart meter observes only net load, while PV generation is estimated from net load and solar-weather features.}
\label{fig:btm_schematic}
\end{figure*}

Energy Internet operation depends on digital visibility of distributed energy resources for trustworthy coordination~\cite{yang2021exploring}, but residential photovoltaic (PV) generation is often hidden behind the customer meter. Fig.~\ref{fig:btm_schematic} illustrates the visibility problem considered in this paper. Most residential customers have a single net meter that reports the difference between consumption and generation rather than the two quantities separately. Recovering the underlying PV generation profile supports distribution planning, load forecasting, demand-response baseline construction, and monitoring of behind-the-meter resources as distributed PV penetration grows~\cite{IEA2025}.

The disaggregation problem is underdetermined: a single net-load value is the difference of two unobserved non-negative quantities, and there are infinitely many consumption-generation pairs consistent with any given net measurement. Practical visibility therefore depends on learned structure, specifically the diurnal and weather-driven shape of PV generation and the comparatively irregular shape of household consumption. This structure is only partially shared across households: PV orientation, capacity, shading, appliance composition, and occupancy can all produce household-specific signatures. Reliance on these household-specific cues can limit transfer to newly connected or newly instrumented customers.

Existing studies report increasingly strong performance from deep sequence models, graph architectures, and generative approaches~\cite{P4, P15, P16, P31, P32, P33, P34}. For system operators, these results are most useful when the error number is tied to the operating setting in which the visibility estimate will be trusted: the same customers next month, new customers in the same distribution region, or a different region with only a few locally instrumented households. A random timestamp split within the same households evaluates interpolation within familiar customer histories, leaving new-customer and future-period deployment risk unmeasured. PV-disaggregation studies would therefore benefit from reporting the operating regime behind the error number.

We use established sequence models as instruments to answer an operational question: how much PV visibility can a distribution operator expect for the customers, periods, and regions on which a model will actually be used, and which levers improve it? We evaluate six sequence models under four deployment regimes that separate household transfer from temporal extrapolation on Ausgrid, with cross-dataset adaptation to Pecan Street. Our findings are:
\begin{itemize}
    \item Known-customer accuracy substantially understates new-customer error. Under identical future conditions on Ausgrid, daylight mean absolute error (MAE) of the best model rises from 0.104~kW on customers seen during development to 0.185~kW on customers outside the training population.
    \item Deployment setting dominates model choice in our experiments. Across six modern sequence models under a fixed development practice, the deployment regime explains 78.6\% of the error variation in the tested grid and model family only 18.0\%.
    \item Limited target supervision substantially reduces transfer error. Under target-domain normalization, fine-tuning an Ausgrid model on five Pecan Street reference households cuts source-model transfer error by approximately 28\%--36\%, with rapidly saturating returns.
    \item The customer history a deployment needs appears use-case dependent. In our sweeps, longer input windows consistently improve estimates for known customers, while new-customer results vary by model.
\end{itemize}

\section{Related Work}\label{sec:related}

Behind-the-meter PV disaggregation has been studied using physical and proxy-based methods, statistical models, and deep learning. Physical and proxy approaches use irradiance reference signals, clear-sky models, or PV capacity estimates to reconstruct generation~\cite{P5, P22, P30}. These methods are attractive because they encode solar physics, but they can require proxy systems, capacity information, or assumptions about orientation and shading that are often unavailable for residential customers. Learned approaches estimate PV generation directly from net-load and auxiliary features. Recent work includes convolutional neural network-bidirectional long short-term memory (CNN-BiLSTM) models~\cite{P4}, graph and graph-recurrent formulations~\cite{P2, P15, P33}, cross-iteration refinement~\cite{P16}, diffusion models~\cite{P31}, and transfer-oriented long short-term memory (LSTM) variants~\cite{P32}.

We assess whether reported errors answer deployment-relevant visibility questions. Table~\ref{tab:lit_review_summary} audits recent PV-disaggregation papers against the regimes defined in Section~\ref{sec:framework}. Many papers report a percentage split without sufficient detail to identify whether households or time periods overlap between training and testing. Where the split is clear, most studies evaluate on seen households or single deployment regimes. As a result, similar reported MAE values may correspond to different operational uses of the same smart-meter analytics pipeline.

Similar concerns have been raised in non-intrusive load monitoring, where held-out-house and cross-dataset transfer are important for scalable deployment~\cite{AD1}. PV disaggregation is affected by the same issue, but with additional physical structure from irradiance and diurnal PV shape. We therefore position the present work as an operational-regime analysis for distribution-grid visibility that complements architectural work in PV disaggregation.

\begin{table}[t]
\caption{Audit of deployment-regime reporting in residential PV disaggregation literature. \cmark{} indicates the regime is used; \unclear{} indicates insufficient household/time-overlap detail; CD denotes cross-dataset transfer.}
\label{tab:lit_review_summary}
\centering
\scriptsize
\setlength{\tabcolsep}{3pt}
\renewcommand{\arraystretch}{1.12}
\begin{tabular}{@{}L{0.08\columnwidth}L{0.31\columnwidth}C{0.07\columnwidth}C{0.07\columnwidth}C{0.07\columnwidth}C{0.07\columnwidth}C{0.07\columnwidth}@{}}
\toprule
\textbf{ID} & \textbf{Method} & \textbf{S1} & \textbf{S2} & \textbf{S3} & \textbf{S4} & \textbf{CD} \\
\midrule
\cite{P2}  & ST-SGCCaps                       & \unclear & \cmark & \na & \na & \na \\
\cite{P4}  & CNN-BiLSTM                       & \na & \cmark & \na & \na & \na \\
\cite{P12} & ANN                              & \na & \na & \cmark & \na & \na \\
\cite{P15} & Sparse attention graph GRU       & \unclear & \unclear & \na & \na & \na \\
\cite{P16} & LSTM GDFM/CDFM                   & \unclear & \unclear & \na & \na & \na \\
\cite{P31} & Conditional DDPM + U-Net         & \na & \na & \cmark & \na & \cmark \\
\cite{P32} & SAM-LSTM / Syn-LSTM transfer     & \unclear & \unclear & \na & \na & \na \\
\cite{P33} & Dynamic graph neural network     & \unclear & \unclear & \na & \na & \na \\
\cite{P34} & Multi-scale temporal features    & \unclear & \unclear & \na & \na & \na \\
\textbf{Ours} & Deployment-regime analysis    & \cmark & \cmark & \cmark & \cmark & \cmark \\
\bottomrule
\end{tabular}
\end{table}

\section{Distribution-Grid Visibility Problem and Deployment Regimes}\label{sec:framework}

\subsection{PV Visibility from Net Measurements}
For each household $i$ and time $t$, the observed net-load signal is
\begin{equation}
    n_{i,t}=c_{i,t}-g_{i,t}, \quad c_{i,t}\geq0,\; g_{i,t}\geq0,
    \label{eq:net_load}
\end{equation}
where $c_{i,t}$ is gross consumption and $g_{i,t}$ is PV generation. The auxiliary solar-weather channels are global horizontal irradiance (GHI), direct normal irradiance (DNI), diffuse horizontal irradiance (DHI), solar zenith angle (SZA), and ambient temperature $T$. For each household, a training sample consists of a length-$W$ window
\begin{equation}
\mathbf{x}_{i,s:s+W-1}=[n,\mathrm{GHI},\mathrm{DNI},\mathrm{DHI},\mathrm{SZA},T]_{i,s:s+W-1}.
\label{eq:input_window}
\end{equation}
The model predicts a corresponding PV-generation sequence,
\begin{equation}
\hat{\mathbf{g}}_{i,s:s+W-1}=f_{\theta}(\mathbf{x}_{i,s:s+W-1}).
\label{eq:model_output}
\end{equation}

At evaluation, the model receives a complete $W$-step data window before estimating PV for timestamps within that window. The reported evaluation is therefore retrospective and non-causal.

We measure PV-visibility error using daylight MAE in denormalized kW,
\begin{equation}
\MAE_{\mathrm{day}}
=
\frac{\sum_{i,t} m_{i,t}\left|\hat g_{i,t}-g_{i,t}\right|}{\sum_{i,t}m_{i,t}},
\quad
m_{i,t}=\mathbb{1}\{\mathrm{SZA}_{i,t}\leq88^\circ\}.
\label{eq:day_mae}
\end{equation}

\subsection{Operational Deployment Regimes}
\begin{figure}[t]
\centering
\includegraphics[width=\columnwidth]{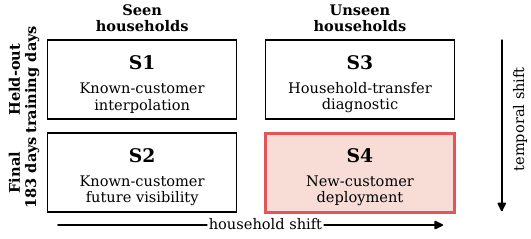}
\caption{Deployment-regime matrix. The regimes separate household transfer from temporal extrapolation. S2 and S4 share the final 183-day test period, isolating the effect of household familiarity.}
\label{fig:regime_matrix}
\end{figure}

Four regimes cross household familiarity and time period, as shown in Fig.~\ref{fig:regime_matrix} and Table~\ref{tab:regimes}. The central operational comparison in this paper is S2 versus S4: both use the same final test period, but S2 represents future visibility for known customers whereas S4 represents future visibility for customers absent from model development.

\begin{table}[t]
\centering
\caption{Deployment regimes. S3 uses held-out households over the pre-test period and is used as a household-transfer diagnostic.}\label{tab:regimes}
\scriptsize
\resizebox{\columnwidth}{!}{%
\begin{tabular}{llll}
\toprule
Regime & Houses & Time period & Purpose \\
\midrule
S1 & Seen & Training-period held-out days & Known-customer interpolation \\
S2 & Seen & Final 183 days & Known-customer future visibility \\
S3 & Unseen & Pre-test period & Household-transfer diagnostic \\
S4 & Unseen & Final 183 days & New-customer future visibility \\
\bottomrule
\end{tabular}%
}
\end{table}

For Ausgrid, 20\% of households are held out by split seed, yielding 239 seen and 60 unseen households. The final 183 days form the future test period for S2 and S4. S1 uses held-out calendar days from seen households in the pre-test period, while S3 uses held-out households in the pre-test period. The validation set is drawn from the pre-test period on seen households only and is used for early stopping and hyperparameter selection. To avoid temporal-window leakage, the split generator rejects any train, validation, or test sample whose input or output window crosses the corresponding split boundary. Held-out households are not used for training, validation, early stopping, or normalization statistics.

\section{Study Design}

\subsection{Datasets and Features}
We evaluate primarily on the Ausgrid Solar Home Electricity Dataset~\cite{AUSGRID}. The original dataset reports consumption and PV generation separately for 300 residential customers in the Ausgrid distribution network around Sydney, Australia, recorded at 30-minute resolution. Our filtered panel contains 299 customers, and the net-load input is synthesized as consumption minus gross PV generation. After filtering, the Ausgrid study window spans January 2011 to June 2013. For cross-dataset transfer, we use Pecan Street~\cite{PECAN}, comprising 17 Austin, Texas households at 30-minute resolution over one year. Solar-weather features are aligned to smart-meter timestamps from NSRDB~\cite{NSRDB}. Evaluation uses historical gridded weather estimates derived from satellite observations and physical models. Operational forecasts are outside the scope of this evaluation. Ausgrid feature normalizers are fit on the training portion of seen households. Pecan Street target-domain normalization uses the 14 non-evaluation households in each split.

\subsection{Visibility Models and Training Contract}
We evaluate six deep sequence models: CNN-BiLSTM~\cite{P4}, temporal convolutional network (TCN)~\cite{TCN}, Temporal Fusion Transformer (TFT)~\cite{TFT}, TimesNet~\cite{TIMESNET}, Mamba~\cite{MAMBA}, and PatchTST~\cite{PATCHTST}. All models receive the same six input channels.

All deep models are trained under a fixed contract. Hyperparameters are selected by a 10-trial random search on the Ausgrid S1 validation split at fixed $W=96$, then frozen and applied unchanged to all regimes. This single-contract setting applies the same visibility-model development practice to every regime, including S4. Neural models use AdamW, cosine learning-rate scheduling with linear warmup, early stopping on validation daylight MAE, non-negativity via softplus output, and zero-clamping at night.

For the main Ausgrid comparison, $W$ is fixed to 96 half-hourly steps (48 hours). The training loss is daylight MAE plus an integrated PV-energy penalty. Because $W$ is fixed in this experiment, the energy term is comparable across regimes and models. For the context-length sweep, where $W$ changes, we use a window-normalized energy loss:
\begin{equation}
L = \MAE_{\mathrm{day}} + \lambda_E\left|\frac{1}{|\mathcal{W}_d|}\sum_{t\in\mathcal{W}_d}\hat{g}_{t} - \frac{1}{|\mathcal{W}_d|}\sum_{t\in\mathcal{W}_d}g_{t}\right|,
\label{eq:normloss}
\end{equation}
where $\mathcal{W}_d$ is the daylight subset of the prediction window. The coefficient $\lambda_E=0.35$ remains fixed from the main validation setting across context lengths. The term discourages window-level PV-energy bias while preventing the regularization magnitude from increasing mechanically with $W$.

Across five Pecan Street splits, three households are evaluated over 2 July--31 December 2018 and $k\in\{1,3,5,10\}$ disjoint reference households are drawn from the remaining 14. The model is fine-tuned on their full 2018 records for 20 epochs (AdamW; 0.1 times the source learning rate; weight decay $10^{-4}$; batch size 32; $W=96$; stride 24; gradient clipping 1.0), requiring approximately 420--4460 updates per split on an AMD RX~7900~XTX. The target-normalized source-model case applies the Ausgrid weights directly; reference-household gradient updates occur in the adaptation cases.

\subsection{Uncertainty and Reproducibility}
Ausgrid results are averaged over five household splits; 95\% BCa intervals are computed by bootstrapping households ($B=1000$). Pecan Street results report mean and 95\% $t$ intervals over five target-household splits (three evaluation households per split; 10 unique), summarizing split-to-split variability. Table~\ref{tab:repro} gives further details.

\begin{table}[t]
\centering
\caption{Reproducibility summary.}
\label{tab:repro}
\scriptsize
\resizebox{\columnwidth}{!}{%
\begin{tabular}{ll}
\toprule
Item & Setting \\
\midrule
Sampling resolution & 30 minutes \\
Input channels & net load, GHI, DNI, DHI, SZA, temperature \\
Primary metric & daylight MAE in denormalized kW \\
Daylight mask & $\mathrm{SZA}\leq88^\circ$ \\
Main context & $W=96$ half-hourly steps \\
Future test period & final 183 days \\
Ausgrid split seeds & five household split seeds \\
Pecan transfer splits & five; three evaluation houses per split \\
Ausgrid uncertainty & BCa bootstrap over households, $B=1000$ \\
Pecan uncertainty & 95\% $t$ interval over five splits \\
Inference setting & offline retrospective, non-causal \\
\bottomrule
\end{tabular}%
}
\end{table}

\section{Results and Discussion}\label{sec:results}

\subsection{Ausgrid: New-Customer Visibility Error is Much Higher}
\begin{figure}[t]
\centering
\includegraphics[width=0.9\columnwidth]{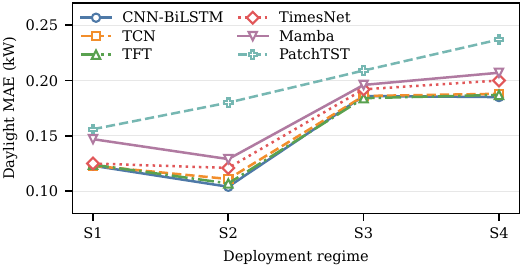}
\caption{Ausgrid daylight MAE across deployment regimes. New-customer regimes produce substantially higher PV-visibility errors than known-customer regimes, while the leading visibility models remain close within each regime.}
\label{fig:ausgrid_mae}
\end{figure}

Table~\ref{tab:ausgrid} and Fig.~\ref{fig:ausgrid_mae} report daylight MAE at fixed $W=96$. The main operational observation is that visibility error changes substantially when the same analytics pipeline is moved from known customers to unseen customers. For example, CNN-BiLSTM increases from 0.104~kW under S2 to 0.185~kW under S4, and TCN increases from 0.111~kW to 0.188~kW.

The seen-household regimes also reveal period-composition effects. S2 is lower than S1 for most models, even though S1 is an interpolation setting, which suggests that seasonal and irradiance differences between held-out pre-test days and the final 183-day period influence absolute MAE. The main deployment comparison therefore centers on S2 versus S4, which share the same final test period and isolate the effect of household familiarity.

\begin{table*}[t]
\centering
\caption{Ausgrid daylight PV-visibility error, measured as MAE (kW) at $W = 96$, mean over five house-split seeds with 95\% BCa bootstrap CIs over households. Bold marks the best performing model per regime.}
\label{tab:ausgrid}
\scriptsize
\setlength{\tabcolsep}{7pt}
\renewcommand{\arraystretch}{1.15}
\begin{tabular}{lcccc}
\toprule
Model & S1 & S2 & S3 & S4 \\
\midrule
\model{CNN-BiLSTM} & \best{0.123} \ci{0.114}{0.132} & \best{0.104} \ci{0.097}{0.112} & 0.186 \ci{0.170}{0.208} & \best{0.185} \ci{0.168}{0.210} \\
\model{TCN}        & 0.123 \ci{0.115}{0.132} & 0.111 \ci{0.103}{0.119} & 0.186 \ci{0.171}{0.203} & 0.188 \ci{0.172}{0.206} \\
\model{TFT}        & 0.124 \ci{0.116}{0.132} & 0.107 \ci{0.100}{0.116} & \best{0.184} \ci{0.168}{0.204} & 0.187 \ci{0.169}{0.212} \\
\model{TimesNet}   & 0.125 \ci{0.117}{0.135} & 0.121 \ci{0.113}{0.131} & 0.192 \ci{0.176}{0.214} & 0.200 \ci{0.182}{0.224} \\
\model{Mamba}      & 0.147 \ci{0.138}{0.158} & 0.129 \ci{0.120}{0.139} & 0.196 \ci{0.183}{0.215} & 0.207 \ci{0.191}{0.227} \\
\model{PatchTST}   & 0.156 \ci{0.146}{0.169} & 0.180 \ci{0.166}{0.201} & 0.209 \ci{0.192}{0.242} & 0.237 \ci{0.223}{0.258} \\
\bottomrule
\end{tabular}
\end{table*}

\begin{figure}[t]
\centering
\includegraphics[width=\columnwidth]{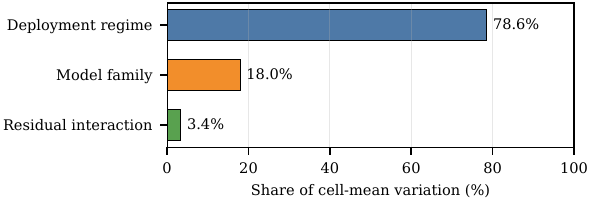}
\caption{Descriptive variance decomposition of the deep-model MAE grid. Under the fixed training contract, deployment regime explains the largest share of cell-mean variation.}
\label{fig:variance_decomp}
\end{figure}

To make this comparison precise, let $y_{a,r}$ denote the mean daylight MAE of model family $a=1,\ldots,A$ under regime $r=1,\ldots,R$, and decompose the cell-mean variation as
\begin{equation}
    y_{a,r}=\mu+\alpha_a+\beta_r+\epsilon_{a,r},
    \label{eq:anova}
\end{equation}
where $\alpha_a$ and $\beta_r$ are model-family and regime main effects. Attributing the total sum of squares of the $A=6$ by $R=4$ grid to these effects in the standard two-way manner, Fig.~\ref{fig:variance_decomp} shows that deployment regime accounts for 78.6\% of the variation, model family for 18.0\%, and residual interaction for 3.4\%. Under this fixed hyperparameter-search budget and training contract, where the PV-visibility estimate is deployed changes the error far more than the choice among the leading visibility models. For a utility procuring or developing PV-visibility analytics, this suggests that the data regime the system will face deserves at least as much scrutiny as model choice.

A rank-level analysis refines this guidance. Ranking the model families by mean MAE within S1 and within S4, the two orderings remain highly correlated (Spearman $\rho_s\approx0.90$, Kendall $\tau_a\approx0.80$). Convenient known-customer tests therefore remain useful for coarse model screening before deployment; deployment-error levels still require regime-matched evaluation. Planning studies, hosting-capacity assessments, and demand-response baselines that rely on PV-visibility estimates for uninstrumented customers should budget against S4-type error, which is nearly twice the known-customer figure in our experiments.

\subsection{Target-Region Fine-Tuning Improves Cross-Region Transfer}
Table~\ref{tab:transfer} reports results for the target-normalized source model and adaptations using $k\in\{1,3,5,10\}$ reference households. Cross-region distribution and seasonal shifts raise error above Ausgrid S4. At $k=5$, every model's source-model MAE falls by 28\%--36\%, with a best MAE of 0.353~kW.

\begin{table*}[t]
\centering
\caption{Ausgrid $\to$ Pecan Street daylight MAE (kW), mean [95\% $t$ interval] over five target-household selection splits. Fine-tuning and evaluation households are disjoint within every split. Bold marks the best model per column.}
\label{tab:transfer}
\scriptsize
\setlength{\tabcolsep}{3.5pt}
\resizebox{\textwidth}{!}{%
\begin{tabular}{lccccc}
\toprule
Model & Source model & $k=1$ & $k=3$ & $k=5$ & $k=10$ \\
\midrule
\model{TCN}         & 0.554 \ci{0.475}{0.634} & \best{0.392 \ci{0.351}{0.434}} & \best{0.373 \ci{0.294}{0.452}} & \best{0.353 \ci{0.306}{0.399}} & \best{0.377 \ci{0.294}{0.460}} \\
\model{CNN-BiLSTM} & \best{0.515 \ci{0.443}{0.587}} & 0.403 \ci{0.369}{0.438} & 0.385 \ci{0.311}{0.459} & 0.367 \ci{0.323}{0.411} & 0.385 \ci{0.306}{0.464} \\
\model{TFT}         & 0.521 \ci{0.454}{0.588} & 0.407 \ci{0.376}{0.437} & 0.389 \ci{0.318}{0.460} & 0.371 \ci{0.336}{0.407} & 0.393 \ci{0.304}{0.483} \\
\model{TimesNet}    & 0.578 \ci{0.490}{0.666} & 0.424 \ci{0.364}{0.484} & 0.402 \ci{0.328}{0.475} & 0.384 \ci{0.328}{0.440} & 0.396 \ci{0.326}{0.466} \\
\model{Mamba}       & 0.589 \ci{0.507}{0.671} & 0.433 \ci{0.388}{0.478} & 0.402 \ci{0.339}{0.465} & 0.393 \ci{0.357}{0.429} & 0.409 \ci{0.342}{0.476} \\
\model{PatchTST}    & 0.713 \ci{0.598}{0.828} & 0.608 \ci{0.543}{0.673} & 0.541 \ci{0.455}{0.627} & 0.510 \ci{0.451}{0.570} & 0.505 \ci{0.422}{0.588} \\
\bottomrule
\end{tabular}%
}
\end{table*}

Gains saturate quickly: moving from $k=1$ to $k=5$ consistently helps, whereas $k=10$ adds little. For TCN, the paired reduction from the source model to $k=5$ is 0.202~kW [95\% $t$ interval: 0.129--0.275~kW]. Larger target datasets would improve precision.

\subsection{Operational Value of Historical Context}\label{subsec:context}

\begin{figure*}[t]
\centering
\includegraphics[width=0.90\textwidth]{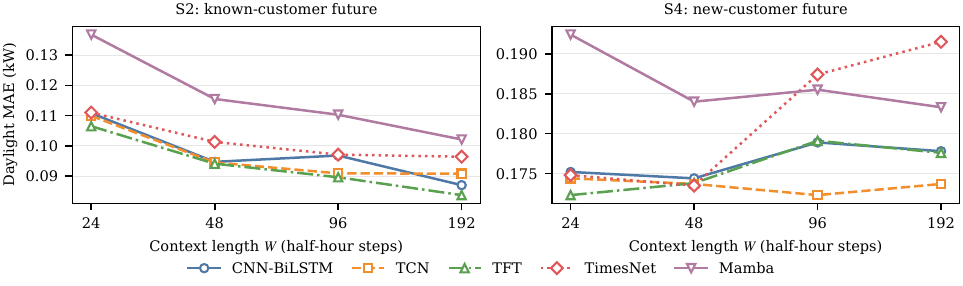}
\caption{Context-length sensitivity under S2 and S4. Long retrospective context consistently improves known-customer temporal extrapolation, whereas new-customer deployment performance is flatter and model-dependent.}
\label{fig:context_length}
\end{figure*}

Many prior PV-disaggregation studies adopt a one-day window because it matches a daily-profile formulation~\cite{P4}, \cite{P15}, \cite{P31}, \cite{P34}. From an operations perspective, context length controls how much recent customer history is required before a visibility estimate can be produced. We evaluate context-length sensitivity across deployment regimes by sweeping $W\in\{24,48,96,192\}$ under the normalized objective in~(\ref{eq:normloss}). To make context lengths comparable, MAE is recomputed on the $W=192$-compatible test timestamp support. This window-normalized analysis covers three household split seeds and the five models shown in Fig.~\ref{fig:context_length}. Its purpose is to compare trends across context lengths; the absolute MAE values differ from Table~\ref{tab:ausgrid} because the sweep uses fewer split seeds, a different loss normalization, and restricted timestamp support.

\begin{table}[t]
\centering
\caption{Summary of the context-length sweep. $\Delta_{192-24}$ is the MAE at $W=192$ minus the MAE at $W=24$. Negative values indicate improvement from longer context.}
\label{tab:context_summary}
\scriptsize
\setlength{\tabcolsep}{4pt}
\resizebox{\columnwidth}{!}{%
\begin{tabular}{lcccc}
\toprule
Model & S2 best $W$ & S2 $\Delta_{192-24}$ & S4 best $W$ & S4 $\Delta_{192-24}$ \\
\midrule
\model{CNN-BiLSTM} & 192 & -0.0238 & 48  &  0.0026 \\
\model{TCN}         & 192 & -0.0191 & 96  & -0.0006 \\
\model{TFT}         & 192 & -0.0228 & 24  &  0.0053 \\
\model{TimesNet}    & 192 & -0.0146 & 48  &  0.0167 \\
\model{Mamba}       & 192 & -0.0346 & 192 & -0.0091 \\
\bottomrule
\end{tabular}%
}
\end{table}

Fig.~\ref{fig:context_length} and Table~\ref{tab:context_summary} show a clear asymmetry. Under S2, all five models achieve their best mean MAE at $W=192$, with improvements from $W=24$ ranging from 0.0146 to 0.0346~kW. Long retrospective context is consistently useful when the household is familiar but the time period is future. Under S4, context preferences vary by model class: CNN-BiLSTM and TimesNet favor $W=48$, TCN favors $W=96$, TFT favors $W=24$, and Mamba favors $W=192$. The S4 differences are also smaller and less stable than the S2 gains.

Long retrospective context likely contains household-specific behavioral and PV-system signatures. These signatures are useful under S2, where the household has been observed during training, and less reliable under S4, where the household is unseen. These results suggest an operational mapping. For retrospective analytics on known customers, such as settlement, demand-response baseline construction, and hosting-capacity studies over an instrumented cohort, supplying several days of history was consistently beneficial in our sweep. New-customer responses varied by model, allowing window size to follow data-availability and latency constraints.

\subsection{Data Quality and Deployment Considerations}
The experiments use cleaned, synchronized 30-minute records. Future work will quantify sensitivity to outages, delays, misalignment, and measurement noise. Online deployment also requires causal windows, inference-time weather, and testing under representative faults.

\section{Conclusion}\label{sec:conclusion}
This paper studied behind-the-meter PV disaggregation as a distribution-grid visibility problem. Visibility error budgets should be regime-matched: error nearly doubled when models served customers outside the training population, while known-customer tests remained useful for coarse model screening. Fine-tuning on a small, household-disjoint target cohort recovered most of the observed adaptation gain, subject to validation on larger datasets. Long input windows improved known-customer estimates; context-length effects for new customers varied by model. PV-visibility results should therefore report deployment regime, leakage-safe splits, and household-level uncertainty.

\bibliography{lib}

@report{IEA2025,
  author      = {{IEA}},
  title       = {Renewables 2025},
  institution = {IEA},
  address     = {Paris},
  year        = {2025},
  url         = {https://www.iea.org/reports/renewables-2025},
  note        = {Licence: CC BY 4.0}
}

@article{P2, title={Behind-the-Meter Load and PV Disaggregation via Deep Spatiotemporal Graph Generative Sparse Coding With Capsule Network}, volume={35}, ISSN={2162-2388}, url={http://dx.doi.org/10.1109/TNNLS.2023.3280078}, DOI={10.1109/tnnls.2023.3280078}, number={10}, journal={IEEE Transactions on Neural Networks and Learning Systems}, publisher={Institute of Electrical and Electronics Engineers (IEEE)}, author={Saffari, Mohsen and Khodayar, Mahdi and Khodayar, Mohammad E. and Shahidehpour, Mohammad}, year={2024}, month=oct, pages={14573–14587} }

@inproceedings{P4, title={Distributed Photovoltaic Power Disaggregation Based on Deep Learning}, url={http://dx.doi.org/10.1109/ACPEE60788.2024.10532594}, DOI={10.1109/acpee60788.2024.10532594}, booktitle={2024 9th Asia Conference on Power and Electrical Engineering (ACPEE)}, publisher={IEEE}, author={Zhang, Junwei and Tan, Zhukui and Tian, Jianmin and Li, Keke and Zhang, Jianfeng}, year={2024}, month=apr, pages={1293–1297} }

@inproceedings{P5, series={e-Energy ’17}, title={SunDance: Black-box Behind-the-Meter Solar Disaggregation}, url={http://dx.doi.org/10.1145/3077839.3077848}, DOI={10.1145/3077839.3077848}, booktitle={Proceedings of the Eighth International Conference on Future Energy Systems}, publisher={ACM}, author={Chen, Dong and Irwin, David}, year={2017}, month=may, pages={45–55}, collection={e-Energy ’17} }

@inproceedings{P12, title={Artificial Neural Network for Disaggregation of Behind-the-Meter Energy Consumption and Generation}, url={http://dx.doi.org/10.1109/INDIN58382.2024.10774215}, DOI={10.1109/indin58382.2024.10774215}, booktitle={2024 IEEE 22nd International Conference on Industrial Informatics (INDIN)}, publisher={IEEE}, author={Khafaf, Nameer Al and McGrath, Brendan and Ali, S.M. Nawazish and Jalili, Mahdi}, year={2024}, month=aug, pages={1–6} }

@inproceedings{P15, title={Sparse Attention Graph Gated Recurrent Unit for Spatiotemporal Behind-The-Meter Load and PV Disaggregation}, url={http://dx.doi.org/10.1109/ECAI61503.2024.10607573}, DOI={10.1109/ecai61503.2024.10607573}, booktitle={2024 16th International Conference on Electronics, Computers and Artificial Intelligence (ECAI)}, publisher={IEEE}, author={Khodayar, Mahdi and Bavil, Ali Farajzadeh and Saffari, Mohsen}, year={2024}, month=jun, pages={1–6} }

@article{P16, title={A Novel Data-Driven Method for Behind-the-Meter Solar Generation Disaggregation With Cross-Iteration Refinement}, volume={13}, ISSN={1949-3061}, url={http://dx.doi.org/10.1109/TSG.2022.3171656}, DOI={10.1109/tsg.2022.3171656}, number={5}, journal={IEEE Transactions on Smart Grid}, publisher={Institute of Electrical and Electronics Engineers (IEEE)}, author={Pan, Keda and Chen, Zhaohua and Lai, Chun Sing and Xie, Changhong and Wang, Dongxiao and Zhao, Zhuoli and Wu, Xiaomei and Tong, Ning and Lei Lai, Loi and Hatziargyriou, Nikos D.}, year={2022}, month=sep, pages={3823–3835} }

@article{P22, title={Capacity and output power estimation approach of individual behind-the-meter distributed photovoltaic system for demand response baseline estimation}, volume={253}, ISSN={0306-2619}, url={http://dx.doi.org/10.1016/j.apenergy.2019.113595}, DOI={10.1016/j.apenergy.2019.113595}, journal={Applied Energy}, publisher={Elsevier BV}, author={Li, Kangping and Wang, Fei and Mi, Zengqiang and Fotuhi-Firuzabad, Mahmoud and Duić, Neven and Wang, Tieqiang}, year={2019}, month=nov, pages={113595} }

@inproceedings{P30, title={Estimation of Behind-the-Meter Solar Generation by Integrating Physical with Statistical Models}, url={http://dx.doi.org/10.1109/SmartGridComm.2019.8909743}, DOI={10.1109/smartgridcomm.2019.8909743}, booktitle={2019 IEEE International Conference on Communications, Control, and Computing Technologies for Smart Grids (SmartGridComm)}, publisher={IEEE}, author={Kabir, Farzana and Yu, Nanpeng and Yao, Weixin and Yang, Rui and Zhang, Yingchen}, year={2019}, month=oct, pages={1–6} }

@article{P31,
title = {Conditional diffusion modeling for probabilistic behind-the-meter PV disaggregation},
journal = {Energy and AI},
volume = {24},
pages = {100692},
year = {2026},
issn = {2666-5468},
doi = {https://doi.org/10.1016/j.egyai.2026.100692},
url = {https://www.sciencedirect.com/science/article/pii/S2666546826000182},
author = {Marc Jené-Vinuesa and Hussain Kazmi and Mònica Aragüés-Peñalba and Andreas Sumper}
}

@article{P32,
title = {Practicality-enhanced behind-the-meter PV power generation disaggregation based on synchronization and transferability fused LSTM framework},
journal = {Energy and AI},
volume = {23},
pages = {100675},
year = {2026},
issn = {2666-5468},
doi = {https://doi.org/10.1016/j.egyai.2026.100675},
url = {https://www.sciencedirect.com/science/article/pii/S2666546826000017},
author = {Chengye Zhang and Huan Long and Zijun Zhang and Jinde Cao}
}

@inproceedings{P33,
author = {Helwani, Lorance and Alskaif, Tarek and Sark, Wilfried},
year = {2025},
month = {10},
pages = {1-5},
title = {Dynamic Graph Neural Networks for Uncertainty-Aware Residential Photovoltaic Power and Load Disaggregation},
doi = {10.1109/ISGTEurope64741.2025.11305351}
}

@INPROCEEDINGS{P34,
  author={Chen, Xiaolu and Huang, Chenghao and Zhang, Yanru and Wang, Hao},
  booktitle={2024 IEEE 8th Conference on Energy Internet and Energy System Integration (EI2)}, 
  title={Season-Independent PV Disaggregation Using Multi-Scale Net Load Temporal Feature Extraction and Weather Factor Fusion}, 
  year={2024},
  volume={},
  number={},
  pages={850-855},
  doi={10.1109/EI264398.2024.10990379}}

@INPROCEEDINGS{AD1,
  author={Murray, David and Stankovic, Lina and Stankovic, Vladimir and Lulic, Srdjan and Sladojevic, Srdjan},
  booktitle={ICASSP 2019 - 2019 IEEE International Conference on Acoustics, Speech and Signal Processing (ICASSP)}, 
  title={Transferability of Neural Network Approaches for Low-rate Energy Disaggregation}, 
  year={2019},
  volume={},
  number={},
  pages={8330-8334},
  doi={10.1109/ICASSP.2019.8682486},
  ISSN={2379-190X},
  month={May},}

@article{NSRDB,
title = {The National Solar Radiation Data Base (NSRDB)},
journal = {Renewable and Sustainable Energy Reviews},
volume = {89},
pages = {51-60},
year = {2018},
issn = {1364-0321},
doi = {https://doi.org/10.1016/j.rser.2018.03.003},
url = {https://www.sciencedirect.com/science/article/pii/S136403211830087X},
author = {Manajit Sengupta and Yu Xie and Anthony Lopez and Aron Habte and Galen Maclaurin and James Shelby}
}

@article{AUSGRID,
author = {Elizabeth L. Ratnam and Steven R. Weller and Christopher M. Kellett and Alan T. Murray},
title = {Residential load and rooftop PV generation: an Australian distribution network dataset},
journal = {International Journal of Sustainable Energy},
volume = {36},
number = {8},
pages = {787--806},
year = {2017},
publisher = {Taylor \& Francis},
doi = {10.1080/14786451.2015.1100196},
URL = { 
        https://doi.org/10.1080/14786451.2015.1100196
},
eprint = { 
        https://doi.org/10.1080/14786451.2015.1100196
}
}

@inproceedings{PECAN,
  title={Pecan street inc.: A test-bed for nilm},
  author={Holcomb, Chris},
  booktitle={International Workshop on Non-Intrusive Load Monitoring, Pittsburgh, PA, USA},
  volume={1},
  year={2012}
}

@misc{PATCHTST,
      title={A Time Series is Worth 64 Words: Long-term Forecasting with Transformers}, 
      author={Yuqi Nie and Nam H. Nguyen and Phanwadee Sinthong and Jayant Kalagnanam},
      year={2023},
      eprint={2211.14730},
      archivePrefix={arXiv},
      primaryClass={cs.LG},
      url={https://arxiv.org/abs/2211.14730}, 
}

@misc{MAMBA,
      title={Mamba: Linear-Time Sequence Modeling with Selective State Spaces}, 
      author={Albert Gu and Tri Dao},
      year={2024},
      eprint={2312.00752},
      archivePrefix={arXiv},
      primaryClass={cs.LG},
      url={https://arxiv.org/abs/2312.00752}, 
}

@misc{TIMESNET,
      title={TimesNet: Temporal 2D-Variation Modeling for General Time Series Analysis}, 
      author={Haixu Wu and Tengge Hu and Yong Liu and Hang Zhou and Jianmin Wang and Mingsheng Long},
      year={2023},
      eprint={2210.02186},
      archivePrefix={arXiv},
      primaryClass={cs.LG},
      url={https://arxiv.org/abs/2210.02186}, 
}

@article{TFT,
title = {Temporal Fusion Transformers for interpretable multi-horizon time series forecasting},
journal = {International Journal of Forecasting},
volume = {37},
number = {4},
pages = {1748-1764},
year = {2021},
issn = {0169-2070},
doi = {https://doi.org/10.1016/j.ijforecast.2021.03.012},
url = {https://www.sciencedirect.com/science/article/pii/S0169207021000637},
author = {Bryan Lim and Sercan \"O. Arık and Nicolas Loeff and Tomas Pfister}
}

@misc{TCN,
      title={An Empirical Evaluation of Generic Convolutional and Recurrent Networks for Sequence Modeling}, 
      author={Shaojie Bai and J. Zico Kolter and Vladlen Koltun},
      year={2018},
      eprint={1803.01271},
      archivePrefix={arXiv},
      primaryClass={cs.LG},
      url={https://arxiv.org/abs/1803.01271}, 
}

@inproceedings{yang2021exploring,
  title={Exploring blockchain for the coordination of distributed energy resources},
  author={Yang, Qing and Wang, Hao},
  booktitle={2021 55th Annual Conference on Information Sciences and Systems (CISS)},
  pages={1--6},
  year={2021},
  organization={IEEE}
}

\end{document}